\documentclass[11pt, a4paper, twocolumn,  copyright, goog]{google}
\usepackage{tikz}
\usepackage{multirow} 
\usetikzlibrary{arrows.meta, decorations.pathreplacing}
\usetikzlibrary{calc}
  \definecolor{skyblue}{HTML}{DCEEFB}
  \definecolor{skyborder}{HTML}{7BAFD4}
  \definecolor{skydark}{HTML}{3A7CA5}
  \definecolor{paneltop}{HTML}{EBF5FB}
  \definecolor{panelbot}{HTML}{E1F0FA}
  \definecolor{verifyteal}{HTML}{2E9E6E}
\usepackage[numbers, sort, square]{natbib}

\definecolor{titleblue}{RGB}{33,76,140}
\definecolor{titlebg}{RGB}{234,242,252}
\definecolor{boxborder}{RGB}{210,218,230}
\definecolor{codebg}{RGB}{248,250,252}
\definecolor{codecomment}{RGB}{90,110,130}
\definecolor{codekeyword}{RGB}{33,76,140}
\definecolor{codestring}{RGB}{160,82,45}
\usepackage{listings}
\usepackage{tcolorbox}
\tcbuselibrary{listings, skins, breakable}
\lstdefinestyle{autoorpython}{
  language=Python,
  basicstyle=\ttfamily\footnotesize,
  keywordstyle=\color{codekeyword}\bfseries,
  commentstyle=\color{codecomment}\itshape,
  stringstyle=\color{codestring},
  backgroundcolor=\color{codebg},
  showstringspaces=false,
  breaklines=true,
  breakatwhitespace=false,
  columns=fullflexible,
  keepspaces=true,
  frame=none,
  aboveskip=0pt,
  belowskip=0pt
}

\newtcolorbox{problemcodebox}[1][]{
  enhanced,
  breakable,
  colback=white,
  colframe=boxborder,
  boxrule=0.55pt,
  arc=1.5mm,
  left=2mm,
  right=2mm,
  top=1.5mm,
  bottom=1.5mm,
  colbacktitle=titlebg,
  coltitle=titleblue,
  coltext=black,
  fonttitle=\bfseries\small,
  title={#1},
  attach boxed title to top left={xshift=0.8mm,yshift*=-\tcboxedtitleheight/2},
  boxed title style={
    frame hidden,
    boxrule=0pt,
    arc=1mm,
    left=2mm,
    right=2mm,
    top=1mm,
    bottom=1mm
  },
  sharp corners=south,
  before upper={\textbf{\small Problem description}\par\smallskip},
  before lower={\textbf{\small Solver code}\par\smallskip}
}
\uselogo{} 
\renewcommand{\copyrightext}{%
  \footerfont
  Published as a conference paper at \textbf{COLM 2026}.\\
  \textcopyright\,\the\year{} Google. All rights reserved%
}
\title{\emph{AutoOR}: Scalably Post-training LLMs to Autoformulate Operations Research Problems}
\correspondingauthor{weishiyan@google.com, sumeet.motwani@eng.ox.ac.uk}

\reportnumber{0001} 

\renewcommand{\today}

\author[1,2]{Sumeet Ramesh Motwani}
\author[1]{Chuan Du}
\author[2]{Aleksander Petrov}
\author[1]{Christopher Davis}
\author[2]{Philip Torr}
\author[1]{Antonio Papania-Davis}
\author[1]{Weishi Yan}
\affil[1]{X, The Moonshot Factory}
\affil[2]{University of Oxford}

\begin{abstract}
Optimization problems are central to decision-making in manufacturing, logistics, scheduling, and other industrial settings. Translating complicated descriptions of these problems into solver-ready formulations requires specialized operations research (OR) expertise, making it hard to scale. We present \textbf{AutoOR}, a scalable synthetic data generation and reinforcement learning pipeline that trains LLMs to autoformulate optimization problems specified in natural language across linear, mixed-integer, and non-linear categories. AutoOR generates verified training data from standard optimization forms and uses solver execution feedback as the reward signal for RL post-training. AutoOR applied to an 8B model achieves \emph{state-of-the-art or competitive} results across six established OR benchmarks, matching significantly larger frontier models. For a non-linear problem class involving physical dynamics, where frontier models score near 0\%, we introduce a curriculum RL strategy that bootstraps from limited initial training data to make this class tractable for post-training. We believe that methods such as AutoOR can significantly accelerate industrial decision-making with AI.
\end{abstract}

\begin{document}

\maketitle

\section{Introduction}

Most real-world decision-making involves modeling and solving optimization problems. In industries such as logistics, manufacturing, energy, and finance, problems often appear as complicated descriptions that must be translated (autoformulated) into mathematical formulations that solvers can compute solutions for \citep{astorga2025autoformulationmathematicaloptimizationmodels}. This translation process requires specialized operations research expertise, making it difficult and expensive to scale. Powerful solvers exist, but are useful only once a correct formulation is in hand \citep{beal2018gekko, bolusani2024scip}.

Recent work has attempted to automate this translation process using LLMs via multi-agent prompting \citep{xiao2024chainofexperts, ahmaditeshnizi2023optimusoptimizationmodelingusing, zhang2025orllmagentautomatingmodelingsolving} and fine-tuning on synthetic data \citep{jiang2025llmoptlearningdefinesolve, yang2025optibenchmeetsresocraticmeasure}. However, public training data for autoformulating optimization problems is scarce and not representative of realistic problems. Existing synthetic datasets suffer from low accuracy or are distilled from stronger models, making them expensive and difficult to scale \citep{lima2025trustworthyoptimizationmodelingagent, Huang_2025}. At the same time, the organizations that most need automated formalization, such as manufacturers or logistics operators, often have domain-specific problem distributions that are private and limited in size \citep{mostajabdaveh2025evaluating}. A practical system must therefore be able to start from a small set of representative problems, scale up verified training signal, and generalize without relying on human annotation or distillation.

We present AutoOR, a scalable synthetic data generation and reinforcement learning pipeline that trains LLMs to autoformulate optimization problems across linear, mixed-integer, and non-linear categories. AutoOR exploits a structural asymmetry in this domain \citep{lu2024mathgeniegeneratingsyntheticdata}. For any optimization category with a known standard form, instantiating the template with valid parameters yields a concrete problem and its correct solver code by construction (see Section \ref{sec:autoor}). Backtranslating this into a natural language description is easy to verify component-wise, whereas the reverse direction (generating correct code from a description) is itself the hard problem \citep{sutton2001verification, song2025mindgapexaminingselfimprovement}. Every valid instantiation yields a verified training pair, and data generation scales with inference compute rather than human effort. This turns autoformulation into a verifiable RL domain where candidate code is executed and checked against ground-truth problem specifications.

\begin{figure*}[t]
  \centering
  \vspace{-10pt}
  \includegraphics[width=0.85\textwidth]{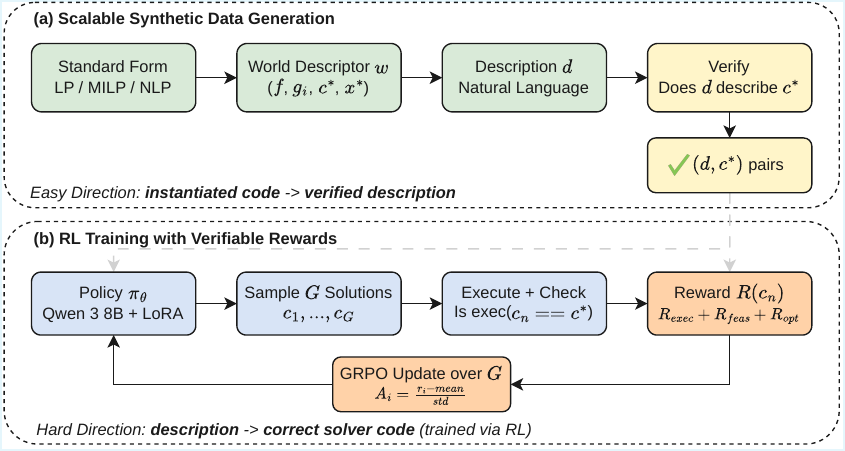}
\caption{\textbf{AutoOR Overview.} \textbf{(a)} We instantiate ground-truth solver code with new parameters, producing synthetic problems that are correct by construction. These are then backtranslated into natural language descriptions. \textbf{(b)} This data enables RL training for the reverse direction: given a description, the policy generates solver code and receives execution feedback as the reward signal.}
  \label{fig:method}
\end{figure*}
Using this pipeline, we post-train Qwen3-8B \citep{yang2025qwen3technicalreport} and show that a small open-source model can match or exceed significantly larger frontier models on OR autoformulation. AutoOR achieves state-of-the-art or competitive results across six established benchmarks (NL4Opt, NL4LP, MAMO-Easy, MAMO-Complex, IndustryOR, ComplexOR) \citep{ramamonjison2023nl4opt, augmentingORModeling, huang2024mamo, Huang_2025, xiao2024chainofexperts} and outperforms baselines on harder evaluations. An 8B model can be deployed at a fraction of frontier API costs and trained on private problem distributions without exposing proprietary data, making automated formalization accessible and scalable.

Nearly all prior work is restricted to linear and mixed-integer programs \citep{Huang_2025, jiang2025llmoptlearningdefinesolve, xiao2025surveyoptimizationmodelingmeets}. Non-linear optimization, which often models physical systems with complex dynamics, is largely unaddressed despite being common in practice \citep{biegler2018new}. Frontier models score near 0\% on some of these problems (Section \ref{sec:frontier}), and cold-start RL fails because all rollouts score zero and the policy gradient vanishes regardless of training duration \citep{yu2025dapoopensourcellmreinforcement}. We introduce a curriculum RL strategy \citep{motwani2025h1bootstrappingllmsreason} that bootstraps from limited initial training data, uses solver syntax as privileged information \citep{qu2026popelearningreasonhard} to break the cold-start barrier, then progressively removes guidance and scales to harder problems, making this problem class tractable for RL post-training. We also take a first step toward multi-turn formalization, training the model to ask clarifying questions when descriptions are incomplete before producing solver code.

\section{Related Work}
\subsection{LLMs for Industrial Optimization}
Translating optimization problems into solver-ready formulations requires OR expertise, 
limiting how broadly optimization can be deployed even though powerful solvers exist once 
a formulation is available \citep{hillier2014introduction}. Several multi-agent and 
prompting-based methods try to automate parts of this pipeline, including Chain-of-Experts 
\citep{xiao2024chainofexperts}, OptiMUS \citep{ahmaditeshnizi2023optimusoptimizationmodelingusing}, 
OR-LLM-Agent \citep{zhang2025orllmagentautomatingmodelingsolving}, SolverLLM 
\citep{li2025solverllmleveragingtesttimescaling}, hierarchical search with SMT-based pruning 
\citep{astorga2025autoformulationmathematicaloptimizationmodels}, and OR-R1 
\citep{ding2025orr1automatingmodelingsolving}. These methods perform no learning from 
experience and are largely restricted to linear and mixed-integer programs. 
\citet{xiao2025surveyoptimizationmodelingmeets} survey existing academic benchmarks, 
find high error rates, and produce corrected versions. We provide additional discussion 
of these methods in Appendix~\ref{app:additional_rw}.

\subsection{Training Methods and Synthetic Data}
A parallel line of work fine-tunes open-source models on synthetic OR data. ORLM \citep{Huang_2025} pioneered this direction with OR-Instruct, a semi-automated synthesis framework that generates natural language--formulation--code triples, achieving competitive performance with GPT-4 when fine-tuning LLaMA-3-8B \citep{grattafiori2024llama3herdmodels} via SFT. However, synthetic data accuracy is low, limiting downstream performance. LLMOPT \citep{jiang2025llmoptlearningdefinesolve} proposed a five-element universal formulation and moved beyond pure SFT by introducing KTO alignment \citep{ethayarajh2024ktomodelalignmentprospect} to reduce hallucinations. ReSocratic \citep{yang2025optibenchmeetsresocraticmeasure} performs reverse (answer-to-question) data synthesis. Step-Opt \citep{wu2025stepoptboostingoptimizationmodeling} augments data through iterative complexity scaling with stepwise validation, and OptMATH \citep{lu2025optmathscalablebidirectionaldata} does bidirectional data synthesis with rejection sampling. AutoOR applies online RL \emph{during training} with solver-verifiable rewards computed against ground-truth problem specifications. Additionally, nearly all prior learning-based methods are restricted to LP and MILP \citep{xiao2025surveyoptimizationmodelingmeets}; we are the first to train models on non-linear optimization problems involving physical dynamics. We follow a scalable backtranslation based synthetic data generation strategy described in Section \ref{backtranslation}.

\subsection{RL for Reasoning and Code Generation}
Group Relative Policy Optimization (GRPO) \citep{shao2024deepseekmathpushinglimitsmathematical} 
eliminates the critic model from PPO \citep{schulman2017proximalpolicyoptimizationalgorithms} 
by sampling groups of outputs and normalizing advantages within each group; DeepSeek-R1 
\citep{Guo_2025} showed that complex reasoning strategies emerge from GRPO with verifiable rewards 
alone, and Dr.\ GRPO \citep{liu2025understandingr1zeroliketrainingcritical} corrects a 
length bias in the objective. More broadly, Reinforcement Learning with Verifiable Rewards 
(RLVR) replaces learned reward models with deterministic programmatic verifiers 
\citep{wen2025reinforcementlearningverifiablerewards}, applied successfully in code generation 
\citep{le2022coderlmasteringcodegeneration, dou2024stepcoderimprovecodegeneration} and formal 
mathematics \citep{hubert2025olympiad, ren2025deepseek}. OR autoformulation fits this 
paradigm naturally: solvers provide deterministic, noise-free verification of feasibility, 
constraint satisfaction, and optimality. AutoOR uses GRPO with these solver-based rewards, 
adapting online RLVR from math and code to operations research.

\section{AutoOR}
\label{sec:autoor}
We formalize the autoformulation task as follows. Given a description $d$ of an optimization problem, the goal is to produce executable solver code $c$, such that running $c$ yields a solution that is feasible and optimal with respect to the underlying problem. We denote the policy as $\pi_\theta$, which generates $c \sim \pi_\theta(\cdot \mid d)$. The central challenge is that $c$ must be a precise formalization of $d$.

\subsection{Optimization Problem Classes}
\label{classes}
We consider three categories of optimization problems with increasing formalization difficulty.

\paragraph{Linear and Mixed-Integer Programming (LP/MILP).} We consider problems of the form:
\begin{equation}
\begin{split}
\min_{\mathbf{x}} \; & \mathbf{c}^\top \mathbf{x} \\
\text{s.t.} \quad & A\mathbf{x} \leq \mathbf{b}, \;
\mathbf{x} \geq \mathbf{0}, \;
x_j \in \mathbb{Z} \; \forall j \in \mathcal{I}.
\end{split}
\label{eq:milp}
\end{equation}
where $\mathbf{x} \in \mathbb{R}^n$ is the vector of decision variables, $\mathbf{c} \in \mathbb{R}^n$ the objective coefficients, and $A \in \mathbb{R}^{m \times n}$, $\mathbf{b} \in \mathbb{R}^m$ encode $m$ linear constraints. The set $\mathcal{I} \subseteq \{1, \ldots, n\}$ indexes integer-constrained variables; when $\mathcal{I} = \emptyset$ this reduces to a linear program, and when $\mathcal{I} = \{1,\ldots,n\}$ to an integer linear program (ILP). We generate LPs across domains such as resource allocation, production planning, and blending, and MILPs covering assignment, scheduling, packing, routing, knapsack, set covering, and network flow problems.

\paragraph{Non-Linear Programming (NLP).} Non-linear problems take the form:
\begin{equation}
\min_{\mathbf{x}} \; f(\mathbf{x})
\quad \text{s.t.} \quad
g_i(\mathbf{x}) \leq 0, \; i = 1, \ldots, m
\label{eq:nlp}
\end{equation}
where $\mathbf{x}$ is the vector of decision variables (e.g., control settings, physical design parameters, or operating points), $f(\mathbf{x})$ is a (potentially non-linear) objective such as energy use, pressure loss, or cost, and each $g_i(\mathbf{x}) \le 0$ encodes a constraint. In our setting, these problems model physical systems governed by non-linear dynamics. For example, pump network optimization involves pressure-flow relationships described by the Darcy-Weisbach equation \citep{westerlund1994optimization}. The model must encode these physical laws correctly in solver code, making this category substantially harder, as the model must correctly encode the governing equations, not just map constraints and data to solver syntax.

\subsection{Synthetic Data and Backtranslation}
\label{backtranslation}
\label{backtranslation}
A key challenge in training models for autoformulation is the availability of high-quality description-code pairs. Existing synthetic datasets suffer from low accuracy and limited diversity \citep{Huang_2025}. The natural approach to generating training data is to produce a description and then corresponding code, but verifying whether the code correctly formalizes an ambiguous description is itself hard. We reverse this direction into a more scalable strategy using the four steps here with a detailed analysis in Appendix \ref{app:synth_data}.

\textbf{Stage 1: Standard form selection.} For a given problem category (LP, MILP, or NLP), we select a standard form template $\mathcal{T}$ that defines the structure of the objective, constraints, and decision variables (see Section \ref{classes}).

\textbf{Stage 2: Instantiation.} We use an LLM to instantiate 
$\mathcal{T}$ with concrete parameters, producing a 
\emph{world descriptor}
\[
w = \bigl(f, \{g_i\}, \mathbf{x}^*, c^*, \text{metadata}\bigr).
\]
Here $f$ and $\{g_i\}$ are the concrete objective and constraints,  $c^*$ is ground-truth solver code in Google OR-Tools that solves  the problem, and $\mathbf{x}^*$ is the solution obtained by executing $c^*$. The metadata records variable names, units,  domains, and structural information (e.g., which variables are binary or represent flows). Because $\mathcal{T}$ defines a valid  standard form, any instantiation that meets our validity criteria  (the problem is feasible, $\mathbf{x}^*$ satisfies all constraints, and the problem meets category-specific complexity thresholds) yields a correct $w$ by construction. The ground-truth code $c^*$ and solution $\mathbf{x}^*$ later serve as the basis for computing rewards during RL training.

\textbf{Stage 3: Description generation.} We use an LLM to produce a natural language description $d$ from $w$. The description uses domain-specific language, expresses constraints implicitly, and may omit details that a practitioner would infer from context. We control the degree of complexity through prompt parameters. For the multi-turn setting (Section~\ref{sec:hitl}), we produce incomplete descriptions $d^{-}$ by deliberately omitting specific information such as constraint coefficients, forcing the model to ask clarifying questions first.

\textbf{Stage 4: Description verification.} We verify each generated description $d$ against $w$ by checking whether $d$ contains the correct data values, all constraints, the objective, and variable definitions. Because $w$ is structured, this can be checked component-wise by an LLM, making verification cheap and scalable. Descriptions that fail are discarded and regenerated. We provide more details about these steps in Appendix \ref{app:synth_data}.

\paragraph{The generation-verification gap.} The direction of this pipeline is what makes it scalable. Generating a correct natural language description from a world descriptor is complicated, but verifying one component-wise is cheap (Stage~4). By going from $\mathcal{T} \to w \to d$ rather than $d \to w$, every valid $w$ that meets our criteria yields a verified training instance, and data generation scales with inference compute rather than human annotation. The same asymmetry holds during RL training. Generating correct solver code from a description is the hard problem we are training the model to solve, but verifying a candidate is straightforward since we retain $w$ and can execute the generated code against it. This makes autoformulation a natural fit for reinforcement learning with verifiable rewards.

\paragraph{Non-linear data generation.} For non-linear problems, standard form templates alone are insufficient because the problems involve domain-specific physical dynamics. We therefore bootstrap from a production-related seed problem (centrifugal pump network optimization; \citealt{westerlund1994optimization}) as a representative example. We extract the underlying physical model and generate synthetic variations by varying system parameters, boundary conditions, and network configuration while preserving the governing equations. We then apply the same backtranslation pipeline to produce training data, allowing us to scale from a handful of problems to a large non-linear problem-specific dataset.

\subsection{RL Training}

We train Qwen 3 8B with online RL using GRPO \citep{shao2024deepseekmathpushinglimitsmathematical}. The model receives a natural language description $d$ and generates solver code $c$. Unlike prior work that relies on supervised fine-tuning \citep{Huang_2025, jiang2025llmoptlearningdefinesolve} or test-time pseudo-label RL \citep{ding2025orr1automatingmodelingsolving}, we use online RL with rewards derived from solver execution against the ground-truth code.

\paragraph{Reward function.} For generated code $c$ with corresponding world descriptor $w$, we define a composite reward:
\begin{equation}
R(c, w) = R_{\text{exec}}(c) + R_{\text{feas}}(c, w) + R_{\text{opt}}(c, w)
\label{eq:reward}
\end{equation}
The components are:
\begin{itemize}
    \item $R_{\text{exec}}(c) \in \{0, \alpha_1\}$: whether the code executes without runtime errors.
    \item $R_{\text{feas}}(c, w) \in \{0, \alpha_2\}$: whether the solution satisfies all constraints in $w$, checked by evaluating each $g_i(\hat{\mathbf{x}}) \leq 0$. Only evaluated if $R_{\text{exec}} > 0$.
    \item $R_{\text{opt}}(c, w) \in \{0, \alpha_3\}$: whether the objective value and variables match the ground-truth optimum from $w$ within a tolerance $\epsilon$. Only evaluated if $R_{\text{feas}} > 0$.
\end{itemize}
The weights $\alpha_1 < \alpha_2 < \alpha_3$ reflect increasing difficulty, with the largest reward reserved for a fully correct formulation. For LP, MILP, and NLP, $R_{\text{opt}}$ checks both the optimal objective value and the assigned decision variable values against the ground truth. A spurious formulation producing both the correct objective value and matching variable assignments across all decision variables is unlikely given the size of the feasible space.

\paragraph{GRPO training.} GRPO \citep{shao2024deepseekmathpushinglimitsmathematical} eliminates the need for a learned value function by estimating advantages from groups of sampled outputs. For each description $d$, we sample $G$ code outputs $\{c_1, \ldots, c_G\}$ from the current policy $\pi_{\theta_\text{old}}$. Each output receives reward $r_i = R(c_i, w)$, and the group-normalized advantage is:
\begin{equation}
\hat{A}_{i,t}
=
\frac{r_i - \mathrm{mean}(\{r_j\}_{j=1}^G)}
     {\mathrm{std}(\{r_j\}_{j=1}^G)}
\label{eq:advantage}
\end{equation}
The policy is updated by maximizing the clipped surrogate objective:
\begin{equation}
\resizebox{0.89\columnwidth}{!}{$\displaystyle
\begin{split}
\mathcal{J}_{\text{GRPO}}(\theta)
&= \mathbb{E}_{d}\!\Bigg[
\frac{1}{G}\sum_{i=1}^{G}\frac{1}{|c_i|}\sum_{t=1}^{|c_i|}\\
&\quad\min\!\Big(
\rho_{i,t}\,\hat{A}_{i,t},\;
\mathrm{clip}\big(\rho_{i,t},1\!-\!\epsilon,1\!+\!\epsilon\big)\hat{A}_{i,t}
\Big)\Bigg]
\end{split}
$}
\label{eq:grpo}
\end{equation}
where $\rho_{i,t}(\theta) = \pi_\theta(c_{i,t} \mid d, c_{i,<t}) / \pi_{\theta_\text{old}}(c_{i,t} \mid d, c_{i,<t})$ is the per-token importance sampling ratio and $\epsilon$ is the clipping parameter. This is well-suited to autoformulation. Our reward (\ref{eq:reward}) is computed through program execution, so no learned reward model is needed. The group-based advantage estimation in (\ref{eq:advantage}) handles the high variance in code generation naturally: within a group, some outputs may fail to execute while others produce near-optimal solutions, and normalization produces useful gradient signal regardless of the absolute reward scale.

\paragraph{Training details.}
We train models with LoRA adapters \citep{hu2021loralowrankadaptationlarge,schulman2025lora} for linear programming, mixed integer programming, and non-linear problems based on our synthetic data generation pipeline. Full hyperparameters are in Appendix~\ref{app:hyperparams}.

\section{Going Beyond the Frontier}
\label{sec:frontier}
\begin{figure*}[t]
  \centering
\includegraphics[width=0.95\textwidth]{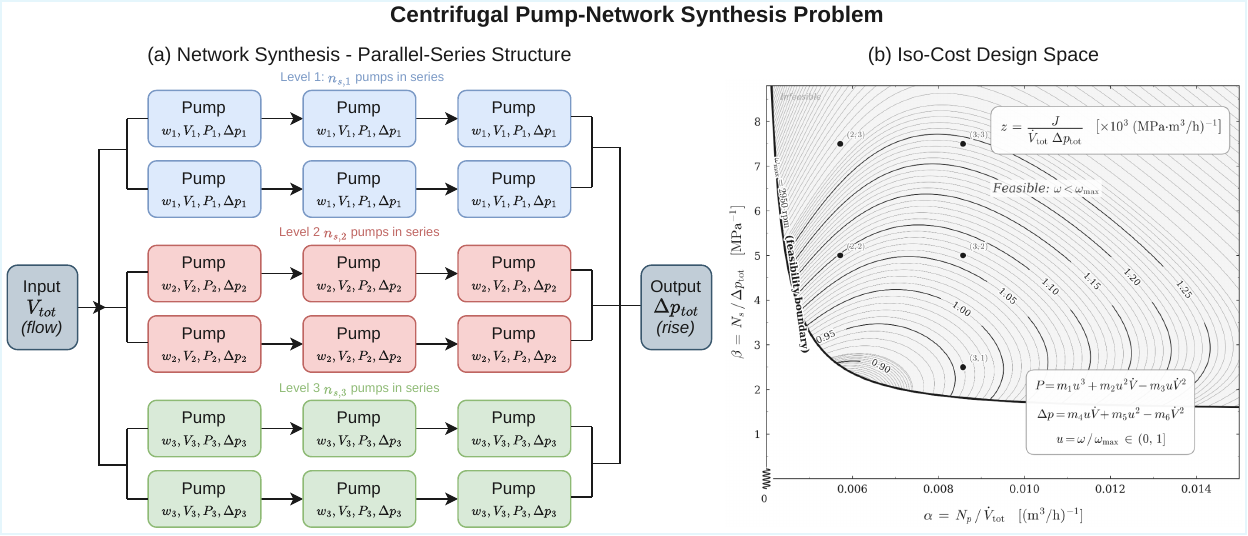}
\caption{\textbf{Pump network synthesis problem} (Section~\ref{sec:frontier}). \textbf{(a)} An $L$-level pump configuration where each level consists of pumps arranged in series-parallel, with level-specific parameters for speed, flow, power, and pressure rise. \textbf{(b)} Normalized objective function contours and feasible region for a single-level optimization instance of this problem.}
  \label{fig:pump}
\end{figure*}


Non-linear optimization problems involving physical dynamics are the hardest category in our pipeline. These problems require correctly encoding the underlying physics, physical constraints, and specialized solver syntax; even minor errors render formulations intractable or unsolvable \citep{tang2024learning}. We demonstrate that AutoOR can learn to autoformulate pump network synthesis problems \citep{floudas2013handbook, westerlund1994optimization}, a practically important MINLP class that combines non-convex pump physics with discrete combinatorial structure. Our results show how AutoOR can be adapted to a specific non-linear problem class on which much larger models fail and where limited training data is available.

\paragraph{The pump network synthesis problem.} Given a total volumetric flowrate $V_{\text{tot}}$ and a target pressure rise $\Delta p_{\text{tot}}$, the goal is to find the least costly configuration of centrifugal pumps arranged across multiple levels. Each level $i$ consists of $n_p^{(i)}$ parallel lines of $n_s^{(i)}$ pumps in series, with a binary variable $z_i$ indicating whether the level is active. Flow is split across levels with fractions $x_i$ satisfying $\sum_i x_i = 1$, and each active level must deliver the full target pressure rise through its series arrangement: $\Delta p_i \cdot n_s^{(i)} = \Delta p_{\text{tot}} \cdot z_i$.

The key difficulty is the pump characteristic curves. Power output and pressure rise are polynomial functions of rotation speed $\omega$ and per-pump flowrate $\dot{v}$:
\begin{equation}
\scalebox{0.98}{$\displaystyle
P = m_1 \!\left(\frac{\omega}{\omega_{\max}}\right)^{\!3}
+ m_2 \!\left(\frac{\omega}{\omega_{\max}}\right)^{\!2} \dot{v}
- m_3 \frac{\omega}{\omega_{\max}} \dot{v}^2
$}
\label{eq:pump_power}
\end{equation}
\begin{equation}
\scalebox{0.98}{$\displaystyle
\Delta p = m_4 \frac{\omega}{\omega_{\max}} \dot{v}
+ m_5 \!\left(\frac{\omega}{\omega_{\max}}\right)^{\!2}
- m_6 \dot{v}^2
$}
\label{eq:pump_pressure}
\end{equation}
where $m_1, \ldots, m_6$ are pump-type-specific coefficients. The objective minimizes total cost:
\begin{equation}
\scalebox{0.98}{$\displaystyle
\min \sum_i \big(C_i + C_i^d \cdot P_i\big) \cdot n_p^{(i)} \cdot n_s^{(i)} \cdot z_i
$}
\label{eq:pump_obj}
\end{equation}
where $C_i$ is the fixed cost and $C_i^d$ is the operating cost coefficient for level $i$. The formulation also includes logical constraints that force all continuous variables to zero when $z_i = 0$, and place upper bounds on power, pressure rise, speed, and flowrate, and the flow-fraction coupling $x_i = \dot{v}_i \cdot n_p^{(i)} / V_{\text{tot}}$. The natural language input describes the number of pumps, cost data, capacity limits, the objective, and the characteristic curves in (\ref{eq:pump_power})--(\ref{eq:pump_pressure}). We generate training data by varying system parameters while preserving the governing equations, scaling from standard forms based on the underlying seed problem (see Sections~\ref{backtranslation} and~\ref{app:synth_data}).



\begin{figure*}[t]
  \centering
\includegraphics[width=1.0\textwidth]{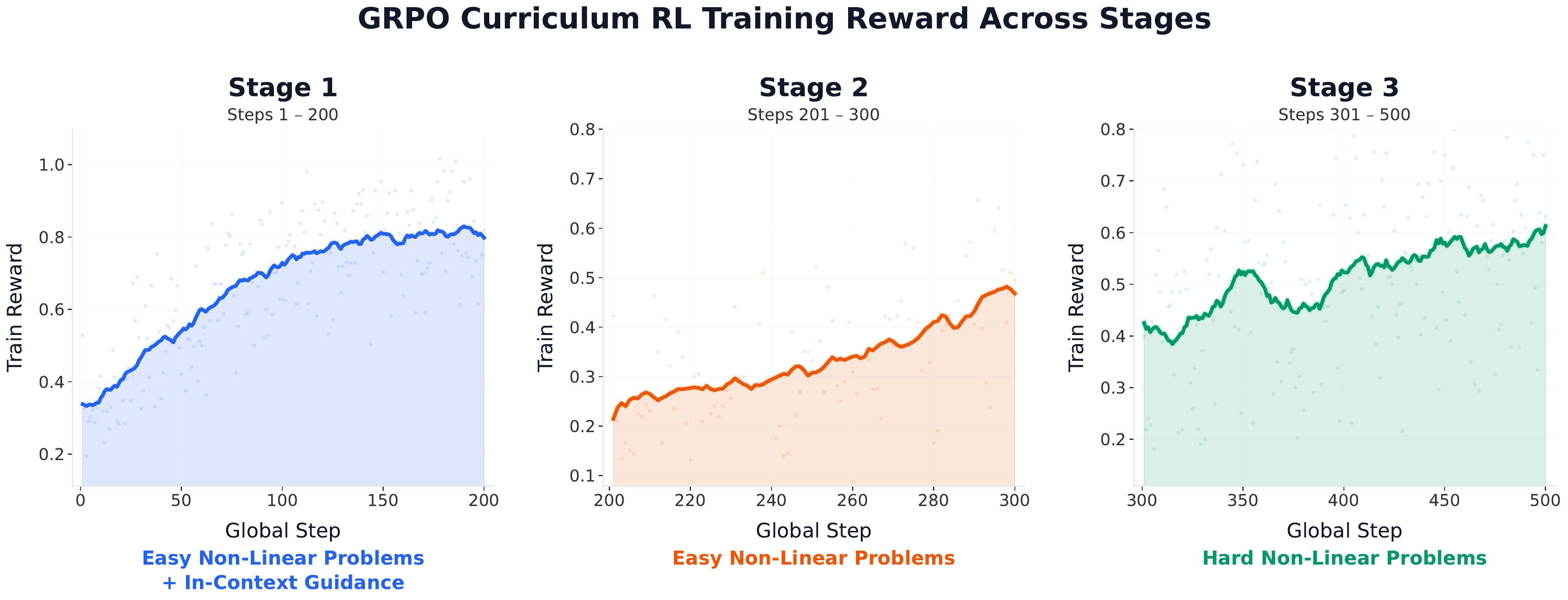}
  \caption{\textbf{Curriculum RL strategy for non-linear optimization.} Phase~1 provides Gekko solver syntax and all constraints as privileged information and trains on easy non-linear problems. Phase~2 removes this guidance and continues training on the same distribution. Phase~3 trains on hard problems from the full non-linear distribution. Each phase produces a policy with sufficient solvability on the next distribution for GRPO to generate learning signal.}
  \label{fig:curriculum}
\end{figure*}

\paragraph{Cold-start problem.} Frontier models fail almost entirely on this problem class. Gemini 3 Pro achieves near-0\% accuracy and untrained Qwen 3 8B scores 0\% even at pass@64. These failures are due to incorrect solver syntax, incorrectly mapping characteristic curves, or not formalizing all the constraints in the system. We note that when Qwen 3 8B is trained directly on these problems, it receives no gradient signal and shows no improvements.

This creates a fundamental problem for RL. We define the solvability of a problem distribution $\mathcal{D}$ under policy $\pi$ as
\begin{equation}
\begin{split}
S(\pi, \mathcal{D}, k)
&= \mathbb{E}_{d \sim \mathcal{D}}\Big[
\mathbf{1}\big\{
\exists\, c_j \in \{c_1, \ldots, c_k\} :\\
&\quad R(c_j, w) > 0
\big\}\Big]
\end{split}
\label{eq:solvability}
\end{equation}
where each $c_j \sim \pi(\cdot \mid d)$. This is the fraction of problems for which at least one of $k$ rollouts obtains positive reward. When $S(\pi_{\text{base}}, \mathcal{D}_{\text{NLP}}, k) \approx 0$, GRPO cannot generate any learning signal: all rollouts within each group receive zero reward and the policy gradient is zero/near-zero regardless of group size or training duration \citep{qu2026popelearningreasonhard}.

\paragraph{Curriculum RL.} We address this with a curriculum that uses privileged information \citep{motwani2025h1bootstrappingllmsreason, qu2026popelearningreasonhard} to guide exploration on problems where the model would otherwise obtain no reward signal (Figure~\ref{fig:curriculum}). Let $\mathcal{D}_{\text{easy}}$ and $\mathcal{D}_{\text{hard}}$ be easy and hard non-linear problem distributions respectively, where difficulty is broadly controlled by the number of pumps, number of levels, pump characteristic coefficients (that change the nonlinear landscape), cost structure, and constraints. The curriculum trains a sequence of policies $\pi_1, \pi_2, \pi_3$ where each phase starts from the previous policy and applies GRPO on a specific distribution. Each phase, for solvability threshold $\tau > 0$ and group size $G$, is viable only if
\begin{equation}
S(\pi_{i-1}, \mathcal{D}_i, G) \geq \tau
\label{eq:curriculum_condition}
\end{equation}

\textbf{Phase 1: Syntax as privileged information.} We provide Gekko \citep{beal2018gekko} solver syntax and mapped out constraints in the prompt and train on $\mathcal{D}_{\text{easy}}$ with GRPO. The privileged information gives the model enough structure to produce executable code on a non-trivial fraction of rollouts, satisfying (\ref{eq:curriculum_condition}) where the base model cannot.

\textbf{Phase 2: Removing guidance.} We remove the syntax information from the prompt and continue training on $\mathcal{D}_{\text{easy}}$. Because $\pi_1$ has internalized the solver syntax and problem structure from Phase~1, it retains sufficient solvability without the privileged information, and RL continues to improve formalization quality.

\textbf{Phase 3: Hard problems.} We train on $\mathcal{D}_{\text{hard}}$, the full non-linear distribution with multiple series-parallel levels and coupled constraints since $S(\pi_2, \mathcal{D}_{\text{hard}}, G) \geq \tau$. We present the training reward curves for these three phases in Figure \ref{fig:curriculum}.

\section{Results}
\begin{table*}[t]
\centering
\small
\definecolor{bestgreen}{HTML}{C8F7C5}
\caption{Main results (pass@1) across established and new benchmarks. 
Accuracy (\%) reported. AutoOR 8B is trained with GRPO on 
synthetic data from our pipeline. For non-linear problems, 
all baseline models score near zero. 
$\dagger$ = introduced in this work. $\diamond$ = sc@3 metric.}
\label{tab:main}
\resizebox{\textwidth}{!}{%
\begin{tabular}{llcccc}
\toprule
Category & Benchmark & Qwen3-8B & Gemini 2.5 Pro 
& Gemini 3 Pro & AutoOR 8B \\
\midrule
\multirow{5}{*}{Linear Programs}
& NL4LP            & 80.56 & 92.59    & 91.67          
& \cellcolor{bestgreen}\textbf{97.22} \\
& NL4OPT           & 59.52 & 69.84    & \textbf{80.95} 
& \cellcolor{bestgreen}78.57 \\
& MAMO-Easy        & 74.00 & 79.00    & 90.00          
& \cellcolor{bestgreen}\textbf{94.00} \\
& MAMO-Complex     & 70.00 & 67.00    & 82.00          
& \cellcolor{bestgreen}\textbf{83.00} \\
& Hard-LP$^\dagger$   & 55.00 & 72.00 & \textbf{85.00}
& \cellcolor{bestgreen}80.00 \\
\midrule
\multirow{3}{*}{\shortstack[l]{Mixed-Integer\\Programs}}
& IndustryOR$^\diamond$       & 52.38 & 74.60    & \textbf{76.19} 
& \cellcolor{bestgreen}69.05 \\
& ComplexOR        & 55.55 & 57.40    & 72.22          
& \cellcolor{bestgreen}\textbf{72.22} \\
& Hard-MIP$^\dagger$  & 64.00 & 84.00  & 92.00         
& \cellcolor{bestgreen}\textbf{94.00} \\
\midrule
\shortstack[l]{Non-Linear}
& Pump-NLP$^\dagger$  & 0.0 & 0.0   & $\approx$0.0  
& \cellcolor{bestgreen}\textbf{48.98} \\
\bottomrule
\end{tabular}%
}
\end{table*}
We evaluate AutoOR 8B (trained on only synthetic data unrelated to these benchmarks) on six established benchmarks and three new benchmarks introduced in this work. Table~\ref{tab:main} compares against the Qwen 3 8B instruct model and frontier models; Table~\ref{tab:learning} compares against published results from learning-based methods. Overall, we use a large range of benchmarks, which allows us to show consistent improvements using our method.

\paragraph{Summary.} Table~\ref{tab:main} shows that AutoOR consistently and substantially improves over the Qwen 3 8B instruct model across all nine benchmarks. AutoOR outperforms Gemini 2.5 Pro on nearly every benchmark and achieves state-of-the-art or competitive results against Gemini 3 Pro, a significantly larger frontier model, across both linear and mixed-integer categories. The improvements are most pronounced on harder benchmarks with longer descriptions and more constraints, suggesting that RL training is particularly effective for problems requiring precise handling of many interacting components. On non-linear problems, the gap is largest: both Qwen 3 8B and Gemini 3 Pro score near 0\% on Pump-NLP, while AutoOR reaches 48.98\%, enabled by the curriculum RL strategy in Section~\ref{sec:frontier}.




\paragraph{Comparison with learning-based methods.} Table~\ref{tab:learning} shows that the largest differences appear on harder benchmarks: AutoOR scores significantly higher than previous learning based methods on Mamo-Complex and MIP problems. These benchmarks have longer descriptions and more complex constraint structures, where the gap between SFT-based methods and online RL is most visible. NL4OPT consists of short, template-like problems where SFT on similar data can be more effective (OptMATH-32B \citep{lu2025optmathscalablebidirectionaldata}); the advantage of RL surfaces on harder problems with more constraints. We also provide an AutoOR-SFT baseline trained on our synthetic data in Table~\ref{tab:learning}.

\begin{table*}[!t]
\centering
\definecolor{bestgreen}{HTML}{C8F7C5}
\caption{Comparison with learning-based methods on established autoformulation benchmarks.
Accuracy (\%) from published results; -- indicates unavailable results.\protect\footnotemark}
\label{tab:learning}
\setlength{\tabcolsep}{1.2\tabcolsep}
\renewcommand{\arraystretch}{0.96}
\begin{tabular}{llcccccc}
\toprule
Method & Size 
& \rotatebox{55}{\scriptsize NL4OPT} 
& \rotatebox{55}{\scriptsize NL4LP} 
& \rotatebox{55}{\scriptsize MAMO-Easy} 
& \rotatebox{55}{\scriptsize MAMO-Cplx} 
& \rotatebox{55}{\scriptsize IndustryOR} 
& \rotatebox{55}{\scriptsize ComplexOR} \\
\midrule
ORLM         & 8B  & 73.8 & 76.4 & 90.4 & 59.5 & 42.9 & 50.0 \\
Step-Opt     & 8B  & 84.5 & --   & 85.3 & 61.6 & 36.4 & -- \\
OptMATH      & 32B & \textbf{95.9} & -- & 89.9 & 54.1 & -- & -- \\
OR-R1\textsubscript{\tiny SFT} & 8B
& 86.0 & 82.9 & 87.0 & 39.9 & 33.0 & 40.7 \\
OR-R1\textsubscript{\tiny RL} & 8B
& 88.3 & 84.6 & 86.1 & 49.9 & 35.3 & 46.3 \\
\midrule
AutoOR\textsubscript{\tiny SFT} & 8B
& 66.7 & 82.9 & 84.0 & 74.0 & 64.3 & 50.0 \\
AutoOR\textsubscript{\tiny RL} & 8B
& \cellcolor{bestgreen}78.6
& \cellcolor{bestgreen}\textbf{97.2}
& \cellcolor{bestgreen}\textbf{94.0}
& \cellcolor{bestgreen}\textbf{83.0}
& \cellcolor{bestgreen}\textbf{69.1}
& \cellcolor{bestgreen}\textbf{72.2} \\
\bottomrule
\end{tabular}
\footnotetext{Base models: ORLM and Step-Opt use LLaMA-3-8B;
OR-R1 and AutoOR use Qwen3-8B; OptMATH uses Qwen2.5-32B.
Numbers are from the original papers.}
\end{table*}

\section{Multi-Turn Training}
\label{sec:hitl}

As LLMs move toward operating as agents in real environments \citep{yao2023reactsynergizingreasoningacting, yang2024sweagentagentcomputerinterfacesenable}, they will rarely receive all necessary information in a single input \citep{ li2025questbenchllmsaskright}. In practice, optimization problems are communicated incrementally and an agent may need to query databases, search documentation, or call simulators over multiple turns to gather sufficient information \citep{laban2025llmslostmultiturnconversation}. We take a first step toward this setting by training AutoOR to gather missing information through a human-in-the-loop setup.

\paragraph{Setup.} We formalize this as a $K$-turn interaction between a policy $\pi_\theta$ and an environment $\mathcal{M}$ that encompasses an oracle with access to the complete problem specification and a solver for evaluating generated code. At each turn, the agent either queries the environment or commits to solver code, receiving outcome reward $R(c, w)$ upon termination. In our current instantiation, $K = 2$. Given a world descriptor $w$, we generate an incomplete description $d^{-}$ by synthetically omitting a subset of information from $w$. Omissions range from explicit data values (e.g., removing a cost coefficient) to less obvious structural information (e.g., removing a constraint that requires domain knowledge to identify as missing). A frontier model with access to $w$ serves as the oracle, simulating a domain expert producing response $o_{i}$. In the first turn, the agent reads $d^{-}$ and produces a clarification query $q_{i}$. In the second turn, it reads the full history and generates solver code $c_{i}$ to autoformulate the problem.


\paragraph{Rewards and credit assignment.} For each incomplete description, we sample $G$ rollouts, each producing a trajectory $\tau_i = (q_i, o_i, c_i)$ with an intermediate reward $R_i^I$ measuring whether the query elicited relevant information in the right format and an outcome reward $R_i^O = R(c_i, w)$ via solver execution. We assign turn-level advantages following multi-turn GRPO \citep{wei2025reinforcingmultiturnreasoningllm}. Let $A_i^I$ and $A_i^O$ be the intermediate and outcome advantages:
\begin{equation}
\hat{A}_{i,1} = A_i^I + \alpha \, A_i^O, \qquad \hat{A}_{i,2} = A_i^O
\label{eq:mt_credit}
\end{equation}
The first turn receives signal from both $R_i^I$ and $R_i^O$; the second is supervised only by $R_i^O$.

\paragraph{Preliminary results.} We evaluate on 100 held-out incomplete LP descriptions with 1--3 omissions. Without the clarification turn (single-turn on $d^{-}$), the model scores 0\%: the missing information is necessary for a correct formulation. In the two-turn setting without multi-turn training, the model achieves 18.50\%, asking questions that are often generic or poorly targeted. After two-turn GRPO training, the model reaches 66.75\%, learning to ask specific, targeted questions (e.g., ``what is the per-unit cost of resource B?'' rather than ``can you provide more details?''). For context, the same model achieves 80\% on these problems in the full information setting (Table~\ref{tab:main}), so multi-turn recovery closes most but not all of the gap introduced by missing information. Extending to $K > 2$ turns with tool calls \citep{singh2025agenticreasoningtoolintegration, feng2025retoolreinforcementlearningstrategic} and training across problem categories are natural next steps.

\vspace{-6pt}
\section{Discussion}
AutoOR demonstrates that a small open-source model can match frontier models on OR autoformulation when trained with online RL on synthetic data generated from standard forms. Our non-linear results are currently restricted to pump network synthesis, and our reward measures correctness but not formulation quality, where equivalent formulations can differ in terms of solve time or formulation size \citep{vielma2015mixed}. In practice, optimization modeling involves multi-step information gathering, tool use, and iterative refinement; our multi-turn setting ($K=2$ turns, simulated oracle, only) is a limited but encouraging step toward this broader agentic setting. We discuss additional limitations in Appendix~\ref{app:limitations}. Extending to additional non-linear domains, optimizing for formulation efficiency, and scaling multi-turn training with tool calls are natural extensions.

More broadly, AutoOR establishes that optimization autoformulation can be turned into a verifiable RL domain through scalable synthetic data generation, and that curriculum RL can make problem classes where even frontier models fail tractable. We believe RL applied to real-world domains like operations research can accelerate industrial decision-making, and methods such as AutoOR are a step toward making expert-level optimization more accessible.
\bibliography{main}
\newpage
\appendix
\onecolumn
\section{Appendix}
\subsection{Additional Related Work}
\label{app:additional_rw}
\paragraph{Prompting and multi-agent methods for OR.}
Chain-of-Experts \citep{xiao2024chainofexperts} orchestrates specialized LLM agents through 
forward thought construction and backward reflection. OptiMUS 
\citep{ahmaditeshnizi2023optimusoptimizationmodelingusing} processes constraints independently 
via connection graphs with retrieval-augmented generation. OR-LLM-Agent 
\citep{zhang2025orllmagentautomatingmodelingsolving} decomposes the pipeline into modeling, 
code generation, and debugging sub-agents built on reasoning LLMs. SolverLLM 
\citep{li2025solverllmleveragingtesttimescaling} applies Monte Carlo Tree Search to 
incrementally construct formulations, and \citet{astorga2025autoformulationmathematicaloptimizationmodels} 
frame autoformulation as a hierarchical search with SMT-based pruning. OR-R1 
\citep{ding2025orr1automatingmodelingsolving} generates multiple candidates and uses majority 
voting to improve performance at inference time. These approaches attempt to scale inference compute 
\citep{snell2024scalingllmtesttimecompute} but perform no learning from experience. AutoOR is an RL post-training method that leads to significant improvements over prompt based methods.

\paragraph{RL for code generation and formal reasoning.}
In code generation, CodeRL \citep{le2022coderlmasteringcodegeneration} pioneered execution-based 
rewards with an actor-critic framework over unit test feedback, and StepCoder 
\citep{dou2024stepcoderimprovecodegeneration} decomposed long code generation into 
curriculum-based subtasks with progressive difficulty, a strategy we adapt for non-linear 
problems. In formal mathematics, AlphaProof \citep{hubert2025olympiad} combined 
AlphaZero-style tree search with Lean verification to reach silver-medal IMO performance, 
and DeepSeek-Prover-V2 \citep{ren2025deepseek} applied GRPO with proof-checker rewards for 
theorem proving.

\subsection{Training Details}
\label{app:hyperparams}

All experiments use Qwen3-8B (which is an instruction tuned model) \citep{yang2025qwen3technicalreport} as the base model with LoRA adapters \citep{hu2021loralowrankadaptationlarge} and bfloat16 precision on NVIDIA A100 GPUs. Single-turn categories (LP, MILP, NLP) are trained with Dr.\ GRPO \citep{liu2025understandingr1zeroliketrainingcritical} via TRL \citep{vonwerra2020trl}; adapters are merged between curriculum phases. Multi-turn training uses Prime-RL with the \texttt{verifiers} library \citep{brown_verifiers_2025}, which employs an importance-ratio masking loss in place of PPO-style clipping and a lower learning rate ($1 \times 10^{-5}$) to preserve the single-turn policy.

Table~\ref{tab:hyper} reports all training configurations. LP and MILP use Google OR-Tools as the target solver; NLP targets Gekko \citep{beal2018gekko} with a higher LoRA rank and temperature to handle the more complex solver code and encourage exploration when solvability is low. NLP curriculum phases correspond to those described in Section~\ref{sec:frontier}: Phase~1 includes privileged solver syntax, Phase~2 removes it, and Phase~3 trains on hard problems. Each phase resumes from the previous checkpoint. All single-turn runs use AdamW ($\beta_1{=}0.9, \beta_2{=}0.99$), weight decay 0.1, cosine LR schedule with warmup ratio 0.1, max gradient norm 0.1, per-device batch size 1 with 8 gradient accumulation steps, and LoRA dropout 0.05 targeting \texttt{q,k,v,o,gate,up,down}\_proj.

\begin{table*}[h]
\centering
\small
\caption{Training configurations by problem category. $G$: rollouts per prompt.}
\label{tab:hyper}
\begin{tabular}{lccccccc}
\toprule
Category & Train & Steps & $G$ & LR & Temp. & LoRA $r$ & Max compl. \\
\midrule
LP & 581 & 400 & 16 & $5 \times 10^{-5}$ & 0.7 & 8 & 14{,}000 \\
MILP & 843 & 600 & 16 & $5 \times 10^{-5}$ & 0.7 & 8 & 12{,}500 \\
\midrule
NLP Phase 1 (w/ syntax, $\mathcal{D}_{\text{easy}}$) & 150 & 200 & 8 & $5 \times 10^{-5}$ & 0.8 & 16 & 12{,}000 \\
NLP Phase 2 (no syntax, $\mathcal{D}_{\text{easy}}$) & 200 & 100 & 8 & $5 \times 10^{-5}$ & 0.8 & 16 & 12{,}000 \\
NLP Phase 3 ($\mathcal{D}_{\text{hard}}$) & 267 & 200 & 8 & $5 \times 10^{-5}$ & 0.8 & 16 & 12{,}000 \\
\midrule
Multi-turn & 262 & 200 & 8 & $1 \times 10^{-5}$ & 0.7 & 16 & 10{,}000 \\
\bottomrule
\end{tabular}
\end{table*}

\paragraph{Reward functions.}
Table~\ref{tab:rewards} summarizes reward components. LP and MILP correctness is checked by comparing optimal objective values against the ground truth, rounded to 2 decimal places (LP additionally compares decision variable values). NLP uses a 2\% relative tolerance on cost with a bonus for matching per-pump power within 5\%. In our setup, outputs are rewarded only if they produce executable solver code whose optimized objective and decision-variable assignments match the ground truth, making reward hacking or non-code based solutions non-viable due to such a large search space of solutions.

\begin{table}[h]
\centering
\small
\caption{Reward components by problem category (additive).}
\label{tab:rewards}
\begin{tabular}{llc}
\toprule
Category & Component & Value \\
\midrule
\multirow{3}{*}{LP / MILP}
  & Code executes without error & 0.1 \\
  & Solution satisfies constraints & 0.1 \\
  & Optimal value + variables match ground truth & 1.0 \\
\midrule
\multirow{3}{*}{NLP}
  & Correct format \& solver initialization & 0.1 \\
  & Solution satisfies constraints & 0.1 \\
  & Cost and configuration within 2\% of ground truth & 1.0 \\
\midrule
\multirow{2}{*}{Multi-turn}
  & Optimal value matches ground truth & 1.0 \\
  & Used clarification query with correct tags & 0.2 \\
\bottomrule
\end{tabular}
\end{table}

\paragraph{Using Online RL.}
AutoOR is an RL post-training method by design: the central contribution is a pipeline that turns autoformulation into a verifiable RL domain. Recent work has established that online RL with outcome-based rewards generalizes to out-of-distribution problems, whereas SFT tends to memorize training data and degrades on unseen variants \citep{chu2025sftmemorizesrlgeneralizes}. This distinction has particularly been shown in in code generation, where \citep{chen2024unlockcorrelationsupervisedfinetuning, wei2025swerladvancingllmreasoning, sancaktar2026deepdivescalingrl} show that RL alleviates overfitting introduced by SFT and improves generalization to unseen programs. In our setting, the evaluation benchmarks for linear and mixed integer programming are disjoint from the training distribution, so the model must generalize rather than only recall. SFT would reduce this to behavior cloning on synthetic solver code, forgoing the exploration that RL provides. Moreover, our domain offers deterministic, noise-free verification through solver execution, making it a natural fit for RLVR \citep{wen2025reinforcementlearningverifiablerewards}. Empirically, the largest gaps between AutoOR and prior SFT-based methods (Table~\ref{tab:learning}) appear on harder benchmarks with longer descriptions and more constraints (MAMO-Complex, IndustryOR, ComplexOR), precisely where generalization matters most and memorization is least useful.

\paragraph{Evaluation Protocol.}
We use Qwen 3 8B and AutoOR-8B with a temperature of 0.7. Gemini 3 Pro and Gemini 2.5 Pro are used with their default temperature. These temperatures are the best suggested configurations from both providers \footnote{\url{https://ai.google.dev/gemini-api/docs/gemini-3} and \url{https://qwen.readthedocs.io/en/latest/getting_started/quickstart.html}} ensuring fair baselines. For each question, all models receive the exact same prompt, and each model produces a single completion evaluated with the same standardized verification scripts and the same maximum completion length. A prediction is counted correct only if it produces executable solver code and passes the benchmark’s correctness check. Our evaluations are across a large range of benchmarks, which allows us to show consistent generalized improvements obtained with our method. The breadth of evaluation across nine benchmarks spanning three problem categories and multiple difficulty levels reduces the likelihood that improvements are attributable to benchmark-specific variance.

\subsection{Synthetic Data Generation Details}
\label{app:synth_data}

\paragraph{Standard forms and code generation.}
For linear programming, we use one standard form (Equation~\ref{eq:milp}) instantiated through five templates that scale the number of decision variables from 4 to 12, with separate splits for continuous LPs solved via GLOP and discrete LPs solved via SCIP. For mixed-integer programming, nine problem families derive from one seed form: assignment, scheduling, packing, routing, network flows, integer linear programs, production planning, knapsack, and set covering. For both LP and MILP, we use Gemini 2.5 Pro with structured outputs\footnote{\url{https://ai.google.dev/gemini-api/docs/structured-output}} to generate concrete parameter values (objective coefficients, constraint bounds, variable relationships) according to a category-specific JSON schema; these values are programmatically inserted into the standard form template to produce ground-truth solver code $c^*$. In principle, these steps require reliable instruction following and structured output generation rather than expert optimization reasoning, and mathematical correctness is guaranteed entirely by the standard form and solver execution. AutoOR substantially outperforms Gemini 2.5 Pro itself on downstream benchmarks (Table~\ref{tab:main}), confirming that the model serves as a data generation tool and that we do not rely on its optimization capabilities. For non-linear programming, we use one seed template (pump network synthesis; Section~\ref{sec:frontier} and derived from \url{https://www.gams.com/latest/gamslib_ml/libhtml/gamslib_pump.html}) and vary difficulty entirely programmatically: the governing equations and constraint structure are fixed, and we control the number of pump types, series and parallel counts, characteristic coefficients, cost structure, and operating conditions without LLM-based instantiation. The ground-truth solution ($\mathbf{x}^*$ and optimal objective value) is solver-produced in all categories: obtained by executing $c^*$, not generated by a language model. We retain only instances where the code executes without error, the solver finds an optimal solution, and no more than one quarter of decision variables take trivial or inactive values at the optimum (zero or at their lower bounds), which would indicate an under-constrained problem. For pump problems, this corresponds to no more than one quarter of pump types being entirely inactive. Approximately 50\% of generated instances pass all code-level checks.

\paragraph{Description generation and verification.}
Descriptions are generated by Gemini 2.5 Pro from the world descriptor $w$. Each generation prompt is randomly paired with one of 8--12 problem scenarios (e.g., manufacturing, logistics, agriculture, energy) and varied paraphrase instructions, so that structurally similar problems produce different natural language representations. The problems themselves differ in objective structure, constraints, number and type of variables, and numerical parameters. Verification with Gemini 2.5 Pro uses five component-wise checks via structured outputs: (1)~all data values for decision variables are present, (2)~all constraints from the code appear in the description, (3)~the objective is correctly stated, (4)~all governing equations and scalar parameters are described, and (5)~the description is self-consistent. Each accepted description must contain sufficient information to recover the unique intended formulation; the verifier checks completeness as well as correctness. Approximately 15\% of generated descriptions pass all five checks. We manually audited 50 accepted pairs across all categories and found no verification errors or missing information. Combined with the $\approx$ 50\% code acceptance rate, 7--8\% of initial generations are used as accepted training instances. This makes the pipeline a way to turn inference compute into verified training data that scales without human annotation and can then be used for reinforcement learning \citep{jayalath2026computeteacherturninginference}.

\paragraph{Uniqueness, tolerance, and generalization.}
We compare objective values and decision variable assignments across all accepted instances within each category and reject duplicates; in practice, collisions are negligible given the size of the parameter space. We also reject rare cases where solver outputs indicate duplicate optima, though such cases were negligible. For LP and MILP, the target solvers (GLOP, SCIP) are deterministic and guarantee global optimality, so for a correctly formulated problem the solution is unique regardless of how the generated code orders variables or constraints; we require match to 2 decimal places during training (Table~\ref{tab:rewards}). For NLP, we use Gekko with the APOPT solver, which solves non-convex MINLPs via branch-and-bound with gradient-based NLP relaxations at each node. Because the continuous subproblems are non-convex, APOPT converges to local optima that can depend on variable ordering, initial point selection, and constraint sequencing in the model. Two mathematically correct formulations of the same pump problem may therefore converge to different locally optimal solutions, so we apply a 2\% relative tolerance on the objective during NLP training (Table~\ref{tab:rewards}). The synthetic training data is generated entirely from clean standard forms and is disjoint from all evaluation benchmarks. This design exploits the fact that optimization problems within a category share a fixed underlying mathematical structure, and RL training over diverse instantiations of this structure enables generalization to held-out problems with different domains, variable names, and constraints.

\subsection{Examples}
We provide examples for all three categories: linear programming, mixed integer programming, and non-linear programming with inputs and associated solver code outputs.

\begin{problemcodebox}[Example: Linear Programming]
A large farming cooperative is planning its operations for the upcoming season to maximize its total profit. The cooperative has 2000 acres of arable land and needs to decide how to allocate this land among three crops: Corn, Soybeans, and Wheat. In addition to crops, they are also considering how many units of cattle to raise, where each unit consists of 10 head. The cooperative's goal is to determine the optimal number of acres for each crop and the number of cattle units to raise. The financial details are as follows: Corn can be sold on the market for \$4 per bushel, and Soybeans for \$10 per bushel. The planting and cultivation costs are \$300 per acre for Corn and \$250 per acre for Soybeans. For Wheat, the cooperative has a long-term contract that guarantees a net profit of \$160 per acre. Raising cattle yields a net profit of \$1000 per unit. The cooperative's operations are subject to several constraints. First, the total available water is 5000 units. Water consumption is 3 units per acre for Corn, 2 units per acre for Soybeans, 1.5 units per acre for Wheat, and 10 units per cattle unit. Second, there are 4000 labor hours available. Labor is required at a rate of 2.5 hours per acre for Corn, 2 hours per acre for Soybeans, 1 hour per acre for Wheat, and 15 hours per cattle unit. The crop yields are 180 bushels per acre for Corn and 50 bushels per acre for Soybeans. A portion of these crops must be used as feed for the cattle; each cattle unit requires 50 bushels of Corn and 20 bushels of Soybeans. The remaining harvest can be sold. Furthermore, due to a government agricultural diversity program, the acreage dedicated to Wheat must be at least 20\% of the total planted acreage. Finally, the cooperative's silo and transportation capacity limits the total amount of Corn and Soybeans sold on the market to a combined maximum of 100,000 bushels. What is the optimal plan for land allocation and cattle raising that maximizes the cooperative's total profit? In your model, please use the following variables: x\_corn\_acres for the acres of Corn, x\_soy\_acres for the acres of Soybeans, x\_wheat\_acres for the acres of Wheat, x\_corn\_sold for the bushels of Corn sold, x\_soy\_sold for the bushels of Soybeans sold, and x\_cattle\_units for the number of cattle units.
\tcblower

\begin{lstlisting}[style=autoorpython]
from ortools.linear_solver import pywraplp
import json

def solve_farming_cooperative_problem():
    """Solves the farming cooperative linear optimization problem."""
    # Create the linear solver with the GLOP backend.
    solver = pywraplp.Solver.CreateSolver('GLOP')
    if not solver:
        return

    # Create the variables.
    infinity = solver.infinity()
    x_corn_acres = solver.NumVar(0, infinity, 'x_corn_acres')
    x_soy_acres = solver.NumVar(0, infinity, 'x_soy_acres')
    x_wheat_acres = solver.NumVar(0, infinity, 'x_wheat_acres')
    x_corn_sold = solver.NumVar(0, infinity, 'x_corn_sold')
    x_soy_sold = solver.NumVar(0, infinity, 'x_soy_sold')
    x_cattle_units = solver.NumVar(0, infinity, 'x_cattle_units')

    # Define the constraints.
    # Total land limit: x_corn_acres + x_soy_acres + x_wheat_acres <= 2000
    solver.Add(x_corn_acres + x_soy_acres + x_wheat_acres <= 2000, 'total_land_limit')

    # Water consumption limit: 3*x_corn_acres + 2*x_soy_acres + 1.5*x_wheat_acres + 10*x_cattle_units <= 5000
    solver.Add(3 * x_corn_acres + 2 * x_soy_acres + 1.5 * x_wheat_acres + 10 * x_cattle_units <= 5000, 'water_consumption_limit')

    # Labor hours limit: 2.5*x_corn_acres + 2*x_soy_acres + x_wheat_acres + 15*x_cattle_units <= 4000
    solver.Add(2.5 * x_corn_acres + 2 * x_soy_acres + x_wheat_acres + 15 * x_cattle_units <= 4000, 'labor_hours_limit')

    # Corn production balance: x_corn_sold + 50*x_cattle_units - 180*x_corn_acres <= 0
    solver.Add(x_corn_sold + 50 * x_cattle_units - 180 * x_corn_acres <= 0, 'corn_production_balance')

    # Soybean production balance: x_soy_sold + 20*x_cattle_units - 50*x_soy_acres <= 0
    solver.Add(x_soy_sold + 20 * x_cattle_units - 50 * x_soy_acres <= 0, 'soybean_production_balance')

    # Crop diversity requirement: -0.2*x_corn_acres - 0.2*x_soy_acres + 0.8*x_wheat_acres >= 0
    solver.Add(-0.2 * x_corn_acres - 0.2 * x_soy_acres + 0.8 * x_wheat_acres >= 0, 'crop_diversity_requirement')

    # Silo storage limit: x_corn_sold + x_soy_sold <= 100000
    solver.Add(x_corn_sold + x_soy_sold <= 100000, 'silo_storage_limit')

    # Define the objective function.
    # Maximize: 4*x_corn_sold + 10*x_soy_sold + 160*x_wheat_acres + 1000*x_cattle_units - 300*x_corn_acres - 250*x_soy_acres
    objective = solver.Objective()
    objective.SetCoefficient(x_corn_sold, 4)
    objective.SetCoefficient(x_soy_sold, 10)
    objective.SetCoefficient(x_wheat_acres, 160)
    objective.SetCoefficient(x_cattle_units, 1000)
    objective.SetCoefficient(x_corn_acres, -300)
    objective.SetCoefficient(x_soy_acres, -250)
    objective.SetMaximization()

    # Solve the system.
    status = solver.Solve()

    # Print the results.
    if status == pywraplp.Solver.OPTIMAL:
        print(f'Optimal value = {solver.Objective().Value()}')
        solution = {
            'x_corn_acres': x_corn_acres.solution_value(),
            'x_soy_acres': x_soy_acres.solution_value(),
            'x_wheat_acres': x_wheat_acres.solution_value(),
            'x_corn_sold': x_corn_sold.solution_value(),
            'x_soy_sold': x_soy_sold.solution_value(),
            'x_cattle_units': x_cattle_units.solution_value()
        }
        print(f'SOLUTION_JSON: {json.dumps(solution)}')
    else:
        print('The problem does not have an optimal solution.')

solve_farming_cooperative_problem()
\end{lstlisting}
\end{problemcodebox}

\begin{problemcodebox}[Example: Mixed Integer Programming]
A logistics manager must determine the optimal cargo allocation across three potential warehouses---North, South, and East---to maximize the net profit, which is calculated as the total profit from stored goods minus the fixed costs associated with opening the facilities. The North warehouse has a storage capacity of 500 m3 and a fixed opening cost of \$1000, the South warehouse has a 400 m3 capacity and an \$800 opening cost, and the East warehouse has a 300 m3 capacity and a \$600 opening cost. Three products can be stored: Widget\_A, which uses 5 m3 and yields \$60 profit per unit; Gadget\_B, which uses 12 m3 and yields \$150 profit per unit; and Gizmo\_C, which uses 8 m3 and yields \$110 profit per unit. Two constraints govern warehouse usage: first, the total volume of products stored in any open warehouse cannot exceed its capacity; and second, any warehouse that is opened must utilize at least 10\% of its available capacity. Additionally, a product mix requirement specifies that the total integer quantity of Gadget\_B units stored across all warehouses must be at least 20\% of the total integer quantity of Widget\_A units stored. The objective is to determine which warehouses to open and the specific integer quantities of each product to store to maximize net profit. Warehouse Data: North (Capacity: 500 m3, Cost: \$1000), South (Capacity: 400 m3, Cost: \$800), East (Capacity: 300 m3, Cost: \$600). Product Data: Widget\_A (Volume: 5 m3, Profit: \$60), Gadget\_B (Volume: 12 m3, Profit: \$150), Gizmo\_C (Volume: 8 m3, Profit: \$110).
\tcblower
\begin{lstlisting}[style=autoorpython]
from ortools.linear_solver import pywraplp
import json

solver = pywraplp.Solver.CreateSolver('SCIP')

# Define variables
open_north = solver.IntVar(0, 1, 'open_north')
open_south = solver.IntVar(0, 1, 'open_south')
open_east = solver.IntVar(0, 1, 'open_east')

widget_a_north = solver.IntVar(0, 100, 'widget_a_north')
widget_a_south = solver.IntVar(0, 80, 'widget_a_south')
widget_a_east = solver.IntVar(0, 60, 'widget_a_east')

gadget_b_north = solver.IntVar(0, 41, 'gadget_b_north')
gadget_b_south = solver.IntVar(0, 33, 'gadget_b_south')
gadget_b_east = solver.IntVar(0, 25, 'gadget_b_east')

gizmo_c_north = solver.IntVar(0, 62, 'gizmo_c_north')
gizmo_c_south = solver.IntVar(0, 50, 'gizmo_c_south')
gizmo_c_east = solver.IntVar(0, 37, 'gizmo_c_east')

# Volume constraints for each warehouse
total_volume_north = 5 * widget_a_north + 12 * gadget_b_north + 8 * gizmo_c_north
solver.Add(total_volume_north <= 500 * open_north)
solver.Add(total_volume_north >= 50 * open_north)

total_volume_south = 5 * widget_a_south + 12 * gadget_b_south + 8 * gizmo_c_south
solver.Add(total_volume_south <= 400 * open_south)
solver.Add(total_volume_south >= 40 * open_south)

total_volume_east = 5 * widget_a_east + 12 * gadget_b_east + 8 * gizmo_c_east
solver.Add(total_volume_east <= 300 * open_east)
solver.Add(total_volume_east >= 30 * open_east)

# Product mix constraint
total_gadget_b = gadget_b_north + gadget_b_south + gadget_b_east
total_widget_a = widget_a_north + widget_a_south + widget_a_east
solver.Add(total_gadget_b >= 0.2 * total_widget_a)

# Objective function
objective = (
    60 * (widget_a_north + widget_a_south + widget_a_east) +
    150 * (gadget_b_north + gadget_b_south + gadget_b_east) +
    110 * (gizmo_c_north + gizmo_c_south + gizmo_c_east) -
    1000 * open_north - 800 * open_south - 600 * open_east
)
solver.Maximize(objective)

# Solve
status = solver.Solve()

if status == pywraplp.Solver.OPTIMAL:
    print(f"Optimal value = {solver.Objective().Value()}")
    variables = {
        'open_north': open_north.solution_value(),
        'open_south': open_south.solution_value(),
        'open_east': open_east.solution_value(),
        'widget_a_north': widget_a_north.solution_value(),
        'widget_a_south': widget_a_south.solution_value(),
        'widget_a_east': widget_a_east.solution_value(),
        'gadget_b_north': gadget_b_north.solution_value(),
        'gadget_b_south': gadget_b_south.solution_value(),
        'gadget_b_east': gadget_b_east.solution_value(),
        'gizmo_c_north': gizmo_c_north.solution_value(),
        'gizmo_c_south': gizmo_c_south.solution_value(),
        'gizmo_c_east': gizmo_c_east.solution_value(),
    }
    print(f"SOLUTION_JSON: {json.dumps(variables)}")
else:
    print("No solution found.")
\end{lstlisting}

\end{problemcodebox}

\begin{problemcodebox}[Example: Non-Linear Pumps]
The objective of this engineering optimization problem is to design a cost-effective centrifugal pump network that satisfies a specific hydraulic requirement. The system must achieve a total target volumetric flow rate of 407.0 units and a total pressure rise of 640.0 units. The design allows for the selection of pump configurations from six distinct pump types. For any selected pump type, the system can utilize a configuration of integer numbers of pumps arranged in series and parallel. The constraints impose a limit of a maximum of 2 pumps in series and a maximum of 2 pumps in parallel for any chosen type. Furthermore, the operational speed of the pumps is variable but must not exceed a maximum limit of 3294.0 RPM. The problem requires determining the optimal mix of active pump types, their specific series and parallel configurations, their operating speeds, and the flow distribution to minimize the total operational and capital cost.
\\\\
The physics of the pumps are modeled using specific algebraic performance curves that relate power and pressure rise to the pump's rotational speed and individual flow rate. For this model, let $w_{\mathrm{ratio}}$ represent the ratio of the operating speed to the maximum speed ($w/3294.0$), and $vdot$ represent the flow rate through a single pump. The power consumption ($P$) for a pump is calculated as coefficient $m_1$ times $w_{\mathrm{ratio}}$ cubed, plus coefficient $m_2$ times $w_{\mathrm{ratio}}$ squared times $vdot$, minus coefficient $m_3$ times $w_{\mathrm{ratio}}$ times $vdot$ squared. The pressure rise ($dp$) across a single pump is calculated as coefficient $m_4$ times $w_{\mathrm{ratio}}$ times $vdot$, plus coefficient $m_5$ times $w_{\mathrm{ratio}}$ squared, minus coefficient $m_6$ times $vdot$ squared. The system enforces flow conservation such that the sum of flow fractions ($x$) across all pump types equals 1 (100\%), where the specific flow allocated to a pump type ($x$ multiplied by the total target flow) must equal the individual pump flow ($vdot$) multiplied by the number of pumps in parallel. Similarly, if a pump type is active, the total system pressure target of 640.0 must equal the pressure rise of a single pump ($dp$) multiplied by the number of pumps in series.
\\\\
The optimization includes binary variables to toggle pump types on or off, along with integer constraints for the number of series and parallel units. If a pump type is active, the chosen power must not exceed its specific maximum power rating ($P_{\max}$), and the configuration must utilize at least one pump in series and one in parallel. The objective function to be minimized is the total cost, calculated as the sum over all pump types of the base cost ($C$) plus the variable power cost ($C_d$ multiplied by the actual Power $P$), all multiplied by the total count of physical pumps (number in series multiplied by number in parallel). The specific parameters for the six available pump types are as follows: Pump Type 0 has a $P_{\max}$ of 165.0, Base Cost ($C$) 8200.0, Power Cost ($C_d$) 1950.0, $m_1$ 21.5, $m_2$ 0.17, $m_3$ 0.00058, $m_4$ 0.72, $m_5$ 345.0, and $m_6$ 0.0125. Pump Type 1 has $P_{\max}$ 185.0, $C$ 9800.0, $C_d$ 1850.0, $m_1$ 28.0, $m_2$ 0.24, $m_3$ 0.00065, $m_4$ 0.88, $m_5$ 410.0, and $m_6$ 0.0190. Pump Type 2 has $P_{\max}$ 210.0, $C$ 11500.0, $C_d$ 1750.0, $m_1$ 36.5, $m_2$ 0.31, $m_3$ 0.00082, $m_4$ 1.05, $m_5$ 465.0, and $m_6$ 0.0260. Pump Type 3 has $P_{\max}$ 155.0, $C$ 7800.0, $C_d$ 2050.0, $m_1$ 19.8, $m_2$ 0.15, $m_3$ 0.00049, $m_4$ 0.68, $m_5$ 325.0, and $m_6$ 0.0110. Pump Type 4 has $P_{\max}$ 230.0, $C$ 13000.0, $C_d$ 1650.0, $m_1$ 42.0, $m_2$ 0.36, $m_3$ 0.00095, $m_4$ 1.15, $m_5$ 520.0, and $m_6$ 0.0320. Finally, Pump Type 5 has $P_{\max}$ 175.0, $C$ 9100.0, $C_d$ 1900.0, $m_1$ 24.5, $m_2$ 0.21, $m_3$ 0.00062, $m_4$ 0.82, $m_5$ 375.0, and $m_6$ 0.0175.

\tcblower

\begin{lstlisting}[style=autoorpython]
from gekko import GEKKO
import numpy as np

# ------------------------------------------------------------------
# 1. Setup Data
# ------------------------------------------------------------------
# Scalars
wmax = 3294.0
Vtot = 407.0
dPtot = 640.0
nsmax = 2
npmax = 2

# Data Dictionary for 6 Pump Types
# Requirement: m5 >= dPtot // 2 (320), Pmax >= 150
pump_data = {
    0: {'Pmax': 165.0, 'C': 8200.0, 'Cd': 1950.0, 'm1': 21.5, 'm2': 0.17, 'm3': 0.00058, 'm4': 0.72, 'm5': 345.0, 'm6': 0.0125},
    1: {'Pmax': 185.0, 'C': 9800.0, 'Cd': 1850.0, 'm1': 28.0, 'm2': 0.24, 'm3': 0.00065, 'm4': 0.88, 'm5': 410.0, 'm6': 0.0190},
    2: {'Pmax': 210.0, 'C': 11500.0, 'Cd': 1750.0, 'm1': 36.5, 'm2': 0.31, 'm3': 0.00082, 'm4': 1.05, 'm5': 465.0, 'm6': 0.0260},
    3: {'Pmax': 155.0, 'C': 7800.0, 'Cd': 2050.0, 'm1': 19.8, 'm2': 0.15, 'm3': 0.00049, 'm4': 0.68, 'm5': 325.0, 'm6': 0.0110},
    4: {'Pmax': 230.0, 'C': 13000.0, 'Cd': 1650.0, 'm1': 42.0, 'm2': 0.36, 'm3': 0.00095, 'm4': 1.15, 'm5': 520.0, 'm6': 0.0320},
    5: {'Pmax': 175.0, 'C': 9100.0, 'Cd': 1900.0, 'm1': 24.5, 'm2': 0.21, 'm3': 0.00062, 'm4': 0.82, 'm5': 375.0, 'm6': 0.0175}
}

N = len(pump_data)

# ------------------------------------------------------------------
# 2. Initialize GEKKO Model
# ------------------------------------------------------------------
m = GEKKO(remote=False)

# ------------------------------------------------------------------
# 3. Variables
# ------------------------------------------------------------------
P = [m.Var(lb=0, ub=pump_data[i]['Pmax'], value=pump_data[i]['Pmax']) for i in range(N)]
w = [m.Var(lb=0, ub=wmax, value=wmax) for i in range(N)]
dp = [m.Var(lb=0, ub=dPtot, value=dPtot) for i in range(N)]
vdot = [m.Var(lb=0, ub=Vtot, value=Vtot) for i in range(N)]
x = [m.Var(lb=0, ub=1.0, value=1.0/N) for i in range(N)]

num_p = [m.Var(lb=0, ub=npmax, integer=True, value=0) for i in range(N)]
num_s = [m.Var(lb=0, ub=nsmax, integer=True, value=0) for i in range(N)]
z = [m.Var(lb=0, ub=1, integer=True, value=0) for i in range(N)]

# ------------------------------------------------------------------
# 4. Equations
# ------------------------------------------------------------------
m.Equation(sum(x) == 1)

for i in range(N):
    d = pump_data[i]
    w_ratio = w[i] / wmax
    
    # Power Output
    m.Equation(P[i] == d['m1'] * w_ratio**3 
                    + d['m2'] * w_ratio**2 * vdot[i] 
                    - d['m3'] * w_ratio * vdot[i]**2)

    # Pressure Rise
    m.Equation(dp[i] == d['m4'] * w_ratio * vdot[i] 
                      + d['m5'] * w_ratio**2 
                      - d['m6'] * vdot[i]**2)
    
    m.Equation(num_p[i] >= z[i])
    m.Equation(num_s[i] >= z[i])

    # Flow and Pressure system connections
    m.Equation(x[i] * Vtot == vdot[i] * num_p[i])
    m.Equation(z[i] * dPtot == dp[i] * num_s[i])

    # Big-M constraints
    m.Equation(w[i] <= wmax * z[i])
    m.Equation(P[i] <= d['Pmax'] * z[i])
    m.Equation(dp[i] <= dPtot * z[i])
    m.Equation(vdot[i] <= Vtot * z[i])
    m.Equation(num_p[i] <= npmax * z[i])
    m.Equation(num_s[i] <= nsmax * z[i])

# ------------------------------------------------------------------
# 5. Objective Function
# ------------------------------------------------------------------
cost_expr = sum((pump_data[i]['C'] + pump_data[i]['Cd'] * P[i]) * num_p[i] * num_s[i] for i in range(N))
m.Minimize(cost_expr)

# ------------------------------------------------------------------
# 6. Solve
# ------------------------------------------------------------------
m.options.SOLVER = 1
m.solve(disp=True)

# ------------------------------------------------------------------
# 7. Print Results
# ------------------------------------------------------------------
print("\n" + "="*60)
print(f"SOLUTION FOUND")
print(f"Total Objective (Cost): ${m.options.OBJFCNVAL:,.2f}")
print("="*60)

print(f"{'Pump':<6} {'On/Off':<8} {'Series':<8} {'Parallel':<10} {'Power (kW)':<12} {'Speed (rpm)':<12} {'Flow Fraction':<15}")

for i in range(N):
    is_on = int(z[i].value[0])
    series = int(num_s[i].value[0])
    parallel = int(num_p[i].value[0])
    p_val = P[i].value[0]
    w_val = w[i].value[0]
    x_val = x[i].value[0]
    
    if is_on:
        print(f"{i+1:<6} {is_on:<8} {series:<8} {parallel:<10} {p_val:<12.2f} {w_val:<12.2f} {x_val:<15.3f}")
    else:
        print(f"{i+1:<6} {'OFF':<8} {'-':<8} {'-':<10} {'0.00':<12} {'0.00':<12} {'0.000':<15}")

# ------------------------------------------------------------------
# The following is the expected output:
# ------------------------------------------------------------------
# SOLUTION FOUND
# Total Objective (Cost): $796,603.26
# Pump   On/Off   Series   Parallel   Power (kW)   Speed (rpm)  Flow Fraction  
# 1      1        2        1          31.54        3294.00      0.201          
# 2      1        2        1          45.02        3294.00      0.235          
# 3      1        2        1          44.79        3047.84      0.188          
# 4      1        2        1          27.77        3294.00      0.168          
# 5      OFF      -        -          0.00         0.00         0.000          
# 6      1        2        1          37.79        3294.00      0.207
\end{lstlisting}
\end{problemcodebox}

\subsection{Evaluations with Additional Models}
We additionally provide an updated comparison with more recent open-weight models. We evaluate three additional models available as of June 2026: DeepSeek-V4-Pro, Llama 4 Maverick, GPT OSS 120B, and Kimi K2.5. We use the same evaluation setup as Table \ref{tab:main}. Table \ref{tab:more} reports these results alongside AutoOR 8B and the baseline Qwen-3-8B model for reference.
\begin{table*}[h]
\centering
\small
\caption{
Additional model evaluations across established and new benchmarks.
Accuracy (\%) is reported. All models are evaluated using the same evaluation
setup as Table~\ref{tab:main}. $\dagger$ = introduced in this work. $\diamond$ = sc@3 metric. $\ddagger$ = Pump-NLP results rounded to nearest percentage point.
}
\label{tab:more}
\resizebox{\textwidth}{!}{%
\begin{tabular}{llcccccc}
\toprule
Category & Benchmark
& Qwen3-8B
& Llama 4 Maverick
& DeepSeek-V4-Pro
& Kimi K2.5
& AutoOR-SFT 8B
& AutoOR-RL 8B \\
\midrule

\multirow{5}{*}{Linear Programs}
& NL4LP
& 80.56
& 97.14
& 88.57
& \textbf{100.00}
& 82.86
& 97.22 \\

& NL4OPT
& 59.52
& 78.57
& \textbf{80.95}
& 78.57
& 66.67
& 78.57 \\

& MAMO-Easy
& 74.00
& 86.00
& 92.00
& 90.00
& 84.00
& \textbf{94.00} \\

& MAMO-Complex
& 70.00
& 28.00
& 34.00
& 34.00
& 74.00
& \textbf{83.00} \\

& Hard-LP$^\dagger$
& 55.00
& 78.00
& \textbf{85.00}
& 74.00
& 26.00
& 80.00 \\

\midrule

\multirow{3}{*}{\shortstack[l]{Mixed-Integer\\Programs}}
& IndustryOR$^\diamond$
& 52.38
& 61.90
& 73.81
& \textbf{76.19}
& 64.29
& 69.05 \\

& ComplexOR
& 55.55
& 61.11
& 72.22
& \textbf{77.78}
& 50.00
& 72.22 \\

& Hard-MIP$^\dagger$
& 64.00
& 58.33
& 68.33
& 55.00
& 45.00
& \textbf{94.00} \\

\midrule

\shortstack[l]{Non-Linear}
& Pump-NLP$^{\dagger,\ddagger}$
& 0.00
& 8.00
& 28.00
& 0.00
& 25.00
& \textbf{49.00} \\

\bottomrule
\end{tabular}%
}
\end{table*}

\subsection{Limitations}
\label{app:limitations}
Our non-linear results are restricted to a single problem class (pump network synthesis), though the curriculum RL strategy is general and we expect it to transfer to other non-linear domains with different governing equations. We also do not address intermediate classes between MILP and fully non-linear programs, such as quadratic programming or second-order cone programs. The model generates code for specific solver APIs (Google OR-Tools for LP/MILP, Gekko for NLP), and generalization to other solvers or solver-agnostic mathematical formulations is not evaluated. Our reward function is coarse: it checks execution, feasibility, and optimality, but does not reward partially correct formulations or penalize formulation inefficiency, treating equivalent formulations that differ dramatically in tractability identically.

All results are shown for Qwen 3 8B, though gains are consistent across all benchmarks tested. More broadly, optimization modeling in practice is a substantially more complex process than the single-turn formalization we focus on, involving iterative refinement, stakeholder communication, model validation, and solver tuning. AutoOR addresses one part of this pipeline, but we show that the method is scalable, and future work on agent post-training and tool use can build on top of it.

\subsection{Acknowledgments}
We thank Chris Hahn, Manu Guere, Bertrand Delorme, Divya Choudhary, Albin L. Jones, Alex Szenicer, Nicholas Dorogy, and M. Pawan Kumar for their feedback and support.

\end{document}